\documentclass{article}

\usepackage{microtype}
\usepackage{graphicx}
\usepackage{subfigure}
\usepackage{booktabs} 
\usepackage{hyperref}
\usepackage{threeparttable}
\usepackage{ascmac}

\usepackage[accepted]{arxiv}

\usepackage{amsmath}
\usepackage{amssymb}
\usepackage{mathtools}
\usepackage{amsthm}
\usepackage[capitalize,noabbrev]{cleveref}
\usepackage{verbatim}

\theoremstyle{plain}

\theoremstyle{definition}

\theoremstyle{remark}

\usepackage[textsize=tiny]{todonotes}
\usepackage{enumitem}
\setlist[itemize]{noitemsep}

\icmltitlerunning{RAP: Retrieval-Augmented Planning with Contextual Memory for Multimodal LLM Agents}

\begin{document}

\twocolumn[
\icmltitle{RAP: Retrieval-Augmented Planning \\with Contextual Memory for Multimodal LLM Agents}



\icmlsetsymbol{equal}{*}

\begin{icmlauthorlist}
\icmlauthor{Tomoyuki Kagaya}{equal,pco}
\icmlauthor{Thong Jing Yuan}{equal,prdcsg}
\icmlauthor{Yuxuan Lou}{equal,nus}
\icmlauthor{Jayashree Karlekar}{prdcsg}
\icmlauthor{Sugiri Pranata}{prdcsg}
\icmlauthor{Akira Kinose}{pco}
\icmlauthor{Koki Oguri}{pco}
\icmlauthor{Felix Wick}{prdcg}
\icmlauthor{Yang You}{nus}
\end{icmlauthorlist}

\icmlaffiliation{pco}{Panasonic Connect Co., Ltd., Japan}
\icmlaffiliation{prdcsg}{Panasonic R\&D Center, Singapore}
\icmlaffiliation{nus}{National University of Singapore, Singapore}
\icmlaffiliation{prdcg}{Panasonic R\&D Center Germany, Germany}

\icmlcorrespondingauthor{Tomoyuki Kagaya}{kagaya.tomoyuki@jp.panasonic.com}
\icmlcorrespondingauthor{Thong Jing Yuan}{jingyuan.thong@sg.panasonic.com}
\icmlcorrespondingauthor{Yuxuan Lou}{yuxuanlou@u.nus.edu}

\icmlkeywords{Machine Learning, ICML}

\vskip 0.3in
]


\printAffiliationsAndNotice{\icmlEqualContribution} 

\begin{abstract}
Owing to recent advancements, Large Language Models (LLMs) can now be deployed as agents for increasingly complex decision-making applications in areas including robotics, gaming, and API integration.  However, reflecting past experiences in current decision-making processes, an innate human behavior, continues to pose significant challenges. Addressing this, we propose Retrieval-Augmented Planning (RAP) framework, designed to dynamically leverage past experiences corresponding to the current situation and context, thereby enhancing agents' planning capabilities. RAP distinguishes itself by being versatile: it excels in both text-only and multimodal environments, making it suitable for a wide range of tasks. Empirical evaluations demonstrate RAP's effectiveness, where it achieves SOTA performance in textual scenarios and notably enhances multimodal LLM agents’ performance for embodied tasks. These results highlight RAP's potential in advancing the functionality and applicability of LLM agents in complex, real-world applications.
\end{abstract}

\begin{figure*}[ht]
\begin{center}
 \includegraphics[width=\textwidth]{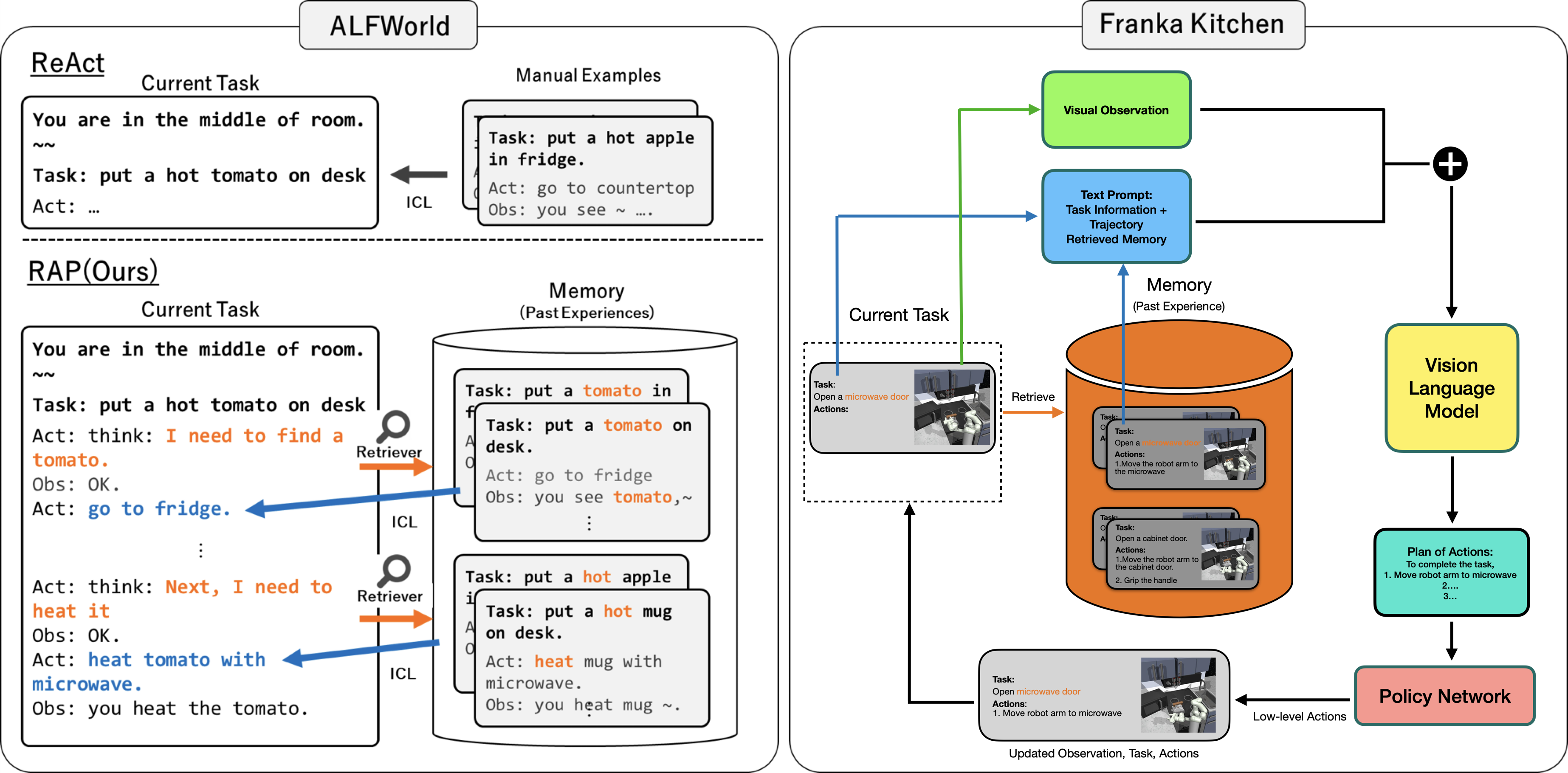}
 \vskip -0.1in
 \caption{Overview of RAP. Our framework stores past experiences and retrieves them based on the current situation. \textbf{Left:} The evaluation process on ALFWorld. ICL stands for in-context learning. \textbf{Right:} The evaluation process on Franka Kitchen.}
 \label{fig:rap}
\end{center}
\vskip -0.2in
\end{figure*}


\section{Introduction}
\label{submission}

Recent research has revealed the high inferential abilities of Large Language Models (LLMs) as agents \cite{wang2023survey, xi2023rise}, indicating their potential application in various areas like decision-making tasks and robotic control.
Previous works such as ReAct \cite{react} have shown that LLMs can generate accurate actions as language agents by iteratively performing actions and reasoning.

Meanwhile, with the rapid development of LLMs, Retrieval-Augmented Generation (RAG) has emerged as a prevalent technique in enhancing the generation capabilities of LLMs. This approach amalgamates external knowledge into the generation process, thus enriching the context and accuracy of generated content. However, integrating external memory to augment LLM agent planning, particularly in diverse environments, poses substantial challenges.
Existing works like Reflexion \cite{reflexion}, which analyzes failure cases, and ExpeL \cite{zhao2023expel}, which extracts insights and enhances learning of language agents, fall short in utilizing comprehensive past information in complex environments.
This highlights a critical gap: the lack of a comprehensive framework for leveraging past experiences in LLM agent planning, limiting their adaptability and applicability in complex, real-world scenarios. 

In this paper, we introduce a novel framework, Retrieval-Augmented Planning (RAP), which embodies a pivotal human ability – to leverage generalized past experiences for current tasks – and apply it to LLM agents.
Our approach involves storing past experiences in memory, retrieving them appropriately based on the similarity with present context including multimodal information, and generating subsequent actions via in-context learning, thereby enhancing the decision-making capacity of language agents.

Central to this framework is the LLMs' ability to perform analogy-making from various abstracted patterns \cite{suvir}. Leveraging this capability, our memory stores both context and action-observation trajectories for each experience.
The approach effectively facilitates deriving correct actions from memory examples within task constraints.
Furthermore, by storing multimodal information in memory and considering it when retrieving past experiences, our approach flexibly utilizes multimodal information with LLMs and Vision-Language Models (VLMs) separately for language agents.

Consequently, our approach proves to be effective for memory utilization by language agents in both decision-making and robotics tasks, in textual and multimodal environments.
Specifically, RAP achieves 33.6\%, 13.0\%, 18.2\%, and 12.7\% gain over ReAct on the ALFWorld, Webshop, Franka Kitchen, and Meta World benchmarks respectively.

To summarize, our contributions are as follows:
\begin{itemize}
\item We propose RAP, a novel framework that enhances LLM agents' planning capacity. It strategically enriches the decision-making process by storing past experiences and intelligently retrieving them based on their similarity to the current situation.
\item RAP is capable of being applied not only in textual environments but also in multimodal embodied tasks, marking it as a pioneering effort in employing memory retrieval techniques for multimodal agents, a first in this domain to our knowledge.
\item We validate the effectiveness of RAP empirically across both textual and multimodal benchmarks. RAP shows significant improvements compared to prior SOTA methods in both types of environments.
\end{itemize}

\begin{figure*}[ht]
\begin{center}
 \includegraphics[width=0.95\textwidth]{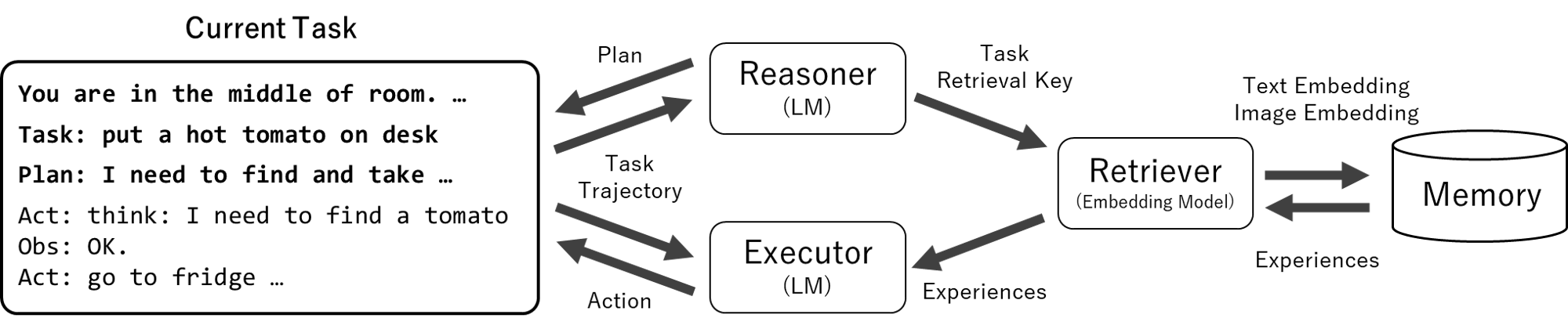}
 \vskip -0.1in
 \caption{RAP Core Components}
 \label{fig:overview}
\end{center}
\vskip -0.2in
\end{figure*}

\section{Related Work}

\subsection{Language Models and Vision-Language Models as Foundations}
Large Language Models (LLMs) such as GPT \cite{gpt4} and the LLaMA series \cite{llama2}, leveraging transformer architecture and self-supervised learning objectives like Next Token Prediction, have excelled in generating coherent, human-like text. These models, pre-trained on extensive text corpora, possess vast linguistic knowledge and reasoning abilities. Extending beyond LLMs, multimodal domains have given rise to Vision-Language Models (VLMs)\cite{yin2023survey} such as LLaVA\cite{Liu2023VisualIT} and CogVLM\cite{Wang2023CogVLMVE}, which integrate textual and visual inputs, exemplified by LLaVA's training on image-caption pairs from the CC3M datasets \cite{sharma-etal-2018-conceptual}. Our work utilizes these foundations to build agents for textual and embodied environments: text-based agents employing LLMs, and embodied agents integrating VLMs for visual perception and action planning. We focus on enhancing these agents' planning capabilities through memory retrieval techniques, enabling them to selectively access and utilize relevant memory for improved sequential decision-making.

\subsection{Language Models as AI Agents}
Recent works have leveraged LLM's anthropomorphic capabilities when building autonomous agents. These agents can be depicted into having 4 key aspects: Profile (agent characteristics), Memory (past information), Planning (future strategies) and Action (execution policies) \cite{autonomous}. A notable example is Chain-of-Thought (CoT) \cite{cot}, where agents are encouraged to mirror human cognitive mechanisms by incorporating reasoning into intermediate steps for complex problem-solving tasks. With a dynamic reasoning process, ReAct \cite{react} interleaves generated actions and environmental states, improving the reasoning ability through action-state synergy. Our work seeks to enhance the ReAct framework by allowing agents to identify specific objects within observations and additionally retrieve relevant aspects of past experiences based on the current context. By doing so, our agent can adaptably receive different experiences at different points in time that are most similar to the situation at hand.  

\subsection{Retrieval-Augmented Generation with Memory}

Among works \cite{madaan2022memory, liu2021makes, su2022selective} that seek to derive better answers from LLMs via memory, RAG \cite{rag} is a notable method that combines retrieval-based mechanisms with generative models. Responses from memory are selected based on similarity and passed into LLMs as additional context to deliver outputs that are creative and contextually-grounded. Building on RAG, Reflexion \cite{reflexion} requires LLMs to self-reflect on unsuccessful tasks for self-improvement in solving tasks over time. ADaPT \cite{prasad2023adapt} further decomposes into sub-tasks and re-executes where necessary. Yet, these works only reflect on trajectories within the task. Hence, these insights are often restricted to each task. Building on Reflexion, ExpeL \cite{zhao2023expel} passes all generated experiences into the LLM to reflect in a text-based manner.
In contrast, our work adopts a different approach by implicitly drawing from a diverse range of experiences from memory without explicitly requiring an additional step of re-tasking the LLM to extract insights. With this approach, our agent can not only efficiently generalize experiences from other successful tasks to solve the current task, but also be flexible enough to extract relevant components from experiences for current task, enhancing the agent's ability to expand its memory from textual to multimodal contexts.

\section{RAP: Retrieval-Augmented Planning}

\begin{figure*}[ht]
\begin{center}
 \includegraphics[width=0.95\textwidth]{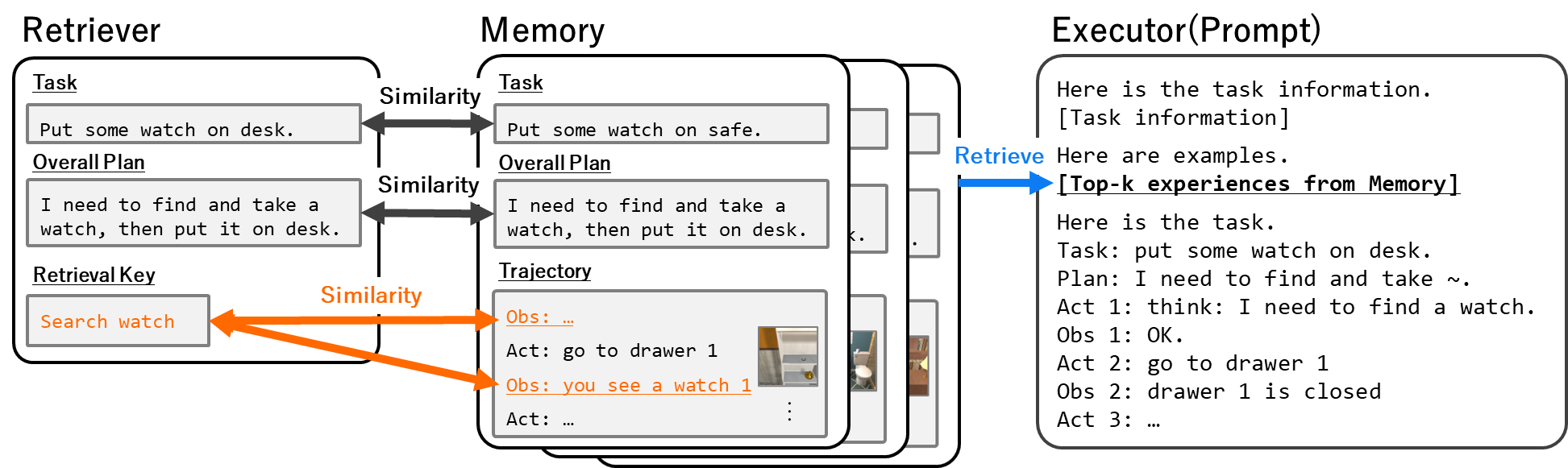}
 \vskip -0.1in
 \caption{Memory-Retrieval in RAP. \textbf{Left to Middle:} The Retriever, calculating similarities with Memory, dynamically switches between action or observation based on the Retrieval Key. This figure illustrates the process of calculating similarity with observation. \textbf{Right:} Executor receives related experiences from memory and utilizes them in the prompt.}
 \label{fig:retriever}
\end{center}
\vskip -0.2in
\end{figure*}

We developed Retrieval-Augmented Planning (RAP), a framework that leverages past experiences to facilitate decision-making according to the current context.
Fig. \ref{fig:overview} provides an overview of the framework, which consists of four core components: Memory, Reasoner, Retriever, and Executor. The specific details of each module will be discussed in sections \ref{memory_section} to \ref{executor_section}.

\subsection{Preliminaries}
In this work, we consider an agent operating in a particular environment and assigned with completing some task T. The agent forms an overall plan $p$, then interacts with the environment over a finite horizon of H timesteps. At each timestep $t \in \{1,2,…H\}$, the agent forms an action plan $p'_t$, selects an action $\alpha_t$ from the action space $A$ and receives an observation $o_t$ from the observation space $O$. The trajectory $\tau = \{\Vec{p'_t}, \Vec{\alpha_t}, \Vec{o_t}\}$ of the agent up to time t consists of the sequences of plans $\Vec{p'_t}=(p'_1,...p'_t)$, actions $\Vec{\alpha_t}=(\alpha_1,...\alpha_t)$ and observations $\Vec{o_t}=(o_1,...o_t)$.

\subsection{Memory}\label{memory_section}
To enable retrieval-augmented planning, we first construct memory databases. The databases contain logs of prior successful task executions. For each log $L_i$ completing a task $T_i$ in $H_i$ steps, we record the task information $T_i$, the overall plan $p_i$ , the trajectory of the agent $\tau_{L_i}$ including plans, actions, and observations sequences.
\begin{equation}
\label{eqn:logs}
L_i = \{T_i, p_i, \tau_{L_i}\}
\end{equation}
\begin{equation}
\label{eqn:trajectory}
\tau_{L_i}= \{\Vec{p'_{L_i}},\Vec{\alpha_{L_i}},\Vec{o_{L_i}}\}
\end{equation}

For textual environments, the observations are textual descriptions of the world state. For multimodal environments, the observations are visual representations - images from a fixed viewpoint camera after each agent action.

The logs are collected by having agents attempt the tasks and saving streams of successful episodes. The episodic logs capture the steps needed to complete the tasks. Storing these examples allows the agents to leverage prior experience when planning for new instances of the tasks.

During interactions with the environment, the agents can selectively retrieve relevant memory samples to make more informed action decisions. For text tasks, the textual logs provide crucial context. For embodiment, prior visual observations reveal outcomes of actions in the space. By retrieving prototypical executions, the agents can plan smarter policies while avoiding past failures. The memory augmentation thus equips the models with vital environmental knowledge for sequential decision-making.

\subsection{Reasoner}

The Reasoner generates overall plans, action plans, and retrieval keys based on the agent's current situation and action trajectory, using LLMs.
Initially, the Reasoner produces the overall plan from the task information.
Based on the task and the overall plan, the Reasoner generates an action plan. Also, in accordance with ReAct, an action or action plan is dynamically generated by LLMs, considering the current task status. If an action plan is generated, a retrieval key is created based on the generated action plan.
For instance, in ALFWorld \cite{alfworld}, if an action plan like "I need to find the watch" is generated, the retrieval key would be "search watch". Hence, the Reasoner enables agents to take into consideration the current situation and context.

\subsection{Retriever}
The Retriever is designed to extract the most relevant memory logs to guide the agent's subsequent actions to complete the current task. This process is shown in Fig. \ref{fig:retriever} (Left).

The similarity score, comparing the current state $S_0$ with log $L_i$, is calculated as a weighted average of the task similarity, overall plan alignment, and retrieval key congruence. 

Let the current agent task be $T_0$, the overall plan be $p_0$, and the retrieval key generated by the Reasoner based on current action plan $p'$ be $k_0$. The similarity score between current state $S_0$ with log $L_i$ is calculated as a weighted average of the similarity score for task, overall plan, and retrieval key.

\begin{align}
    \begin{multlined}
    \label{eqn:overall-similarity}
        Score(S_0, L_i)  =  w_t \cdot sim(T_0, T_{Li}) \\ 
        + w_p \cdot sim(p_0, p_{L_i}) +  w_k \cdot sim(k_0, \tau_{L_i})
    \end{multlined}
\end{align}

Each component's similarity score is determined using cosine similarity of their feature representations. For text data, the representations are derived using sentence-transformers \cite{reimers2019sentencebert}. For images, the representations are generated with a CLIP-based Vision Transformer.

The similarity score between the retrieval key and the log trajectory is adaptive based on environment type and retrieval key type. In multimodal environments, the retrieval key corresponds to agents' current visual observation. Thus, the retrieval-key similarity score is the score between current and logged visual trajectory observations as in equation (\ref{eqn:visual-similarity}).
\begin{equation}
\label{eqn:visual-similarity}
sim(k_0, \tau_{L_i})=max(cos\_sim(k_0^{visual}, o_j))\text{, for } o_j \in \Vec{o_{L_i}}
\end{equation}

In textual environments, the retrieval-key similarity score is adaptive based on key type. In scenarios involving the retrieval key for searching or locating objects, the similarity score is calculated between the retrieval key and the logged textual trajectory observations, as in equation (\ref{eqn:text-similarity}).
\begin{equation}
\label{eqn:text-similarity}
sim(k_0, \tau_{L_i})=max(cos\_sim(k_0^{text}, o_j))\text{, for } o_j  \in \Vec{o_{L_i}}
\end{equation}

For the case retrieval-key is interacting with object action planning, it is an action similarity score, as in equation (\ref{eqn:action-similarity}).
\begin{equation}
\label{eqn:action-similarity}
sim(k_0, \tau_{L_i})=max(cos\_sim(k_0^{text}, \alpha_j))\text{, for } \alpha_j \in \Vec{\alpha_{L_i}}
\end{equation}

Furthermore, component weights are adaptively calibrated based on the environment. In environments with a constrained task space, task similarity is assigned a higher weight, For example, in Franka Kitchen environment which has only 5 tasks. We only retrieve logs with same task type.

For each retrieved experience, only a window of trajectory centered around the most similar action is passed to the agent. This allows agents to focus on actions most similar to current task, rather than full trajectories that may instead create additional noise to the agent. 

In summary, our meticulously-crafted retrieval method efficiently identifies the most pertinent logs by calculating a weighted similarity score that takes into account various aspects including task information, overall planning, and retrieval key. This process ensures that the most relevant and contextually appropriate logs are selected from a vast repository of memory logs. Once these logs are retrieved, they serve as an invaluable resource for the large language model serving as Executor.

\subsection{Executor}\label{executor_section}

The Executor receives past experiences from the Retriever and generates the next action by utilizing these experiences through in-context learning. This process is illustrated in Fig. \ref{fig:retriever} (Right).
By presenting the past experience aligned with the current context as a prompt, it enables accurate decision-making for the next action, mirroring the process humans leverage past experiences for future actions.
Additionally, the length of the current task trajectory is used in the same way as past experiences, utilizing only a constant number of new trajectories. This encourages effective analogy-making from experiences through in-context learning in LLMs.

\begin{algorithm}[tb]
   \caption{Retrieval-Augmented Planning}
   \label{alg:rap}
\begin{algorithmic}
   \STATE {\bfseries Initialize:} Trajectory $\tau$, $\text{Reasoner}_{LLM}$, $\text{Retriever}$, $\text{Executor}_{LLM}$, $\text{Memory}$
   \STATE {\bfseries Input:} Task $T$
   \STATE $p_o$ = Reasoner($T$) // Generate overall plan
   \STATE $e$ = Retriever(Memory, $T$) // Retrieve experiences
   \WHILE{reward is not Success and $t < \text{max steps}$}
     \STATE // Generate action or action plan
     \STATE $a_t$ = Executor($T$, $p_o$, $e$, $\tau$) 
     \IF{action plan}
     \STATE $k$ = Reasoner($a_t$) // Generate retrieval key
     \STATE $e$ = Retriever(Memory, $T$, $k$) // Retrieve experiences
     \ELSE
     \STATE reward, $o_t$ = Env($a_t$) // Input action to environment
     \ENDIF
     \STATE $\tau$ $\leftarrow$ $a_t, o_t$ // Update trajectory
     \STATE $t = t+1$ // Increment
   \ENDWHILE
\end{algorithmic}
\end{algorithm}

\section{Experiments}

\begin{table*}[t]
 \caption{ALFWorld task-specific success rate(\%).}
 \label{alfworld-table}
 \vskip 0.15in
 \begin{center}
 \begin{small}
 \begin{threeparttable}
  \begin{tabular}{c | c | c c c c c c | c}
  \toprule
  Method(d$_{max}$=3) & Model & Pick & Clean & Heat & Cool & Look & Pick2 & All \\
  \midrule
  Act    & GPT-3.5 & 66.7 & 51.6 & 73.9 & 61.9 & 38.9 & 17.6 & 53.7 \\
  ReAct    & GPT-3.5 & 50.0 & 41.9 & 73.9 & 66.7 & 55.6 & 23.5 & 52.2 \\
  Reflexion    & GPT-3.5 & 75.0 & 77.4 & 65.2 & 76.2 & 83.3 & \underline{70.6} & 74.6 \\
  ADaPT\tnote{*}    & GPT-3.5 & 87.5 & 80.6 & 60.8 & 76.2 & 61.1 & 52.9 & 71.6 \\
  RAP(Ours)    & GPT-3.5 & \textbf{95.8} & \underline{87.1} & \underline{78.3} & \textbf{90.5} & \underline{88.9} & \underline{70.6} & \underline{85.8} \\
  RAP$_{train}$(Ours)    & GPT-3.5 & \textbf{95.8} & \textbf{100.0} & \textbf{82.6} & \underline{85.7} & \textbf{100.0} & \textbf{76.5} & \textbf{91.0} \\
  \midrule
  ReAct    & GPT-4 & 83.3 & 71.0 & 95.7 & 81.0 & \textbf{100.0} & \textbf{94.1} & 85.8 \\
  RAP(Ours)    & GPT-4 & \textbf{95.8} & \textbf{90.3} & \textbf{100.0} & \textbf{95.2} & \textbf{100.0} & 88.2 & \textbf{94.8} \\
  \midrule
  ReAct    & Llama2-13b & 29.2 & 41.9 & 34.8 & 52.4 & 38.9 & \textbf{17.6} & 36.6 \\
  RAP(Ours)    & Llama2-13b & \textbf{62.5} & \textbf{61.3} & \textbf{56.5} & \textbf{61.9} & \textbf{44.4} & \textbf{17.6} & \textbf{53.0} \\
  \bottomrule
  \end{tabular}
  \begin{tablenotes}\footnotesize
   \item[*] We use the performance reported by \cite{prasad2023adapt}
  \end{tablenotes}
 \end{threeparttable}
\end{small}
\end{center}
\vskip -0.1in
\end{table*}

To validate the effectiveness of our framework in various environments, we performed evaluations on four benchmarks. These include the text-based multi-step tasks in ALFWorld \cite{alfworld} and Webshop \cite{yao2022webshop}, and robotics tasks, FrankaKitchen \cite{frankakitchen} and Meta-World \cite{yu2021metaworld}, which are multimodal environments with texts and images.

\subsection{Textual Environments}

\subsubsection{ALFWorld}\label{alfworld_experiment}

ALFWorld \cite{alfworld} is a synthetic text-based game that challenges an agent to solve multi-step tasks in a variety of interactive environments based on TextWorld \cite{textworld}. Following ReAct, we evaluated an agent in 134 unseen games, including six types of tasks: Pick, Clean, Heat, Cool, Look, and Pick2.
In this environment, agents are required to accomplish complex tasks using text-based actions in a simulated household providing textual and image feedback.
Following previous works \cite{react, reflexion, prasad2023adapt}, RAP runs recursively until it reaches a depth (trial) of 3.
In RAP$_{train}$, we use 1000 tasks from the provided training set, and run recursively with memory from successful tasks both from the training set and previous trials.
Additionally, we use task information including the task type for the Retriever. During action-based similarity calculation, it extracts four experiences and ten actions from both before and after the most similar action. Meanwhile, during observation-based similarity calculation, it uses eight experiences and five actions from both before and after the most similar action.

We conducted evaluations using three models: GPT-3.5 \cite{gpt3}, GPT-4 \cite{gpt4}, and Llama2-13b \cite{llama2}.
In Table \ref{alfworld-table}, our experiments with GPT-3.5 show that RAP(85.8\%) and RAP$_{train}$(91.0\%) achieve a significantly higher success rate compared to previous works such as ReAct(52.2\%), Reflexion(74.6\%), and ADaPT(71.6\%).
Further, RAP is also effective even with high-performance models like GPT-4, and locally-operated models such as Llama2-13b, thus illustrating the efficacy of our method across various LLMs.

\begin{figure}[t]
\hspace{-100cm}
\begin{center}
 \includegraphics[width=\linewidth-2cm]{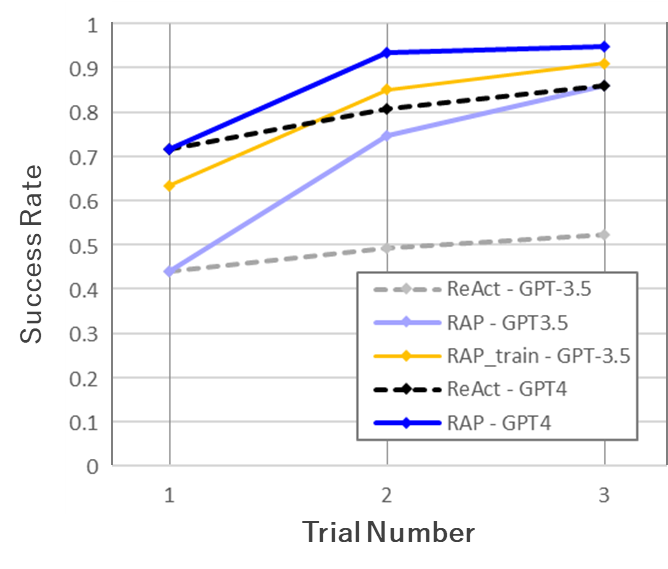}
 \vskip -0.1in
 \caption{Success Rate of ALFWorld with trial number}
 \label{fig:result_alfworld}
\end{center}
\vskip -0.2in
\end{figure}

Figure \ref{fig:result_alfworld} highlights the progression of improvements to success rate over three trials, where RAP shows higher success rates compared to ReAct, thus indicating the effective utility of successful experiences from other tasks in memory. In addition, RAP with GPT-3.5 achieves performance equivalent to ReAct with GPT-4. Also, RAP with memory built from the training set via GPT-3.5 surpasses ReAct with GPT-4.

\subsubsection{WebShop}

WebShop \cite{yao2022webshop} is a web application that simulates online shopping, where agents are required to select products for purchase based on a given user instruction. WebShop contains a total of 1.18M real-world products featured on Amazon, and comprises a wide variety of structured and unstructured texts. Following Reflexion \cite{reflexion} and ADaPT \cite{prasad2023adapt}, we evaluated an agent across 100 instructions. For each instruction, agents are required to reason and select a desired product that is most aligned to the given instruction based on observations returned by the web application, and perform additional precise interactions with the portal to navigate through the web application such as searching or clicking buttons. Such interactions are performed in a text-based manner where the agent issues a textual command into the web application. Following previous studies, we allow the agent to run recursively until it reaches a depth (trial) of 3. 

During our evaluation, we ran the initial trial with a ReAct agent and formulated the memory database based on successful tasks. The memory database would be further expanded for subsequent trials based on successful tasks in the preceding trials. Here, successful tasks are counted as those with a reward of 1. In addition, during retrieval of actions in memory, the Retriever extracts three experiences and five actions from before and after the most similar action.

Moreover, unlike other environments where objects are generalizable across different tasks, WebShop has an additional unique feature where actions in each task are highly dependent on the scenario outlined in that task. As such, apart from the correlation between the current reasoning and trajectories in memory, our agent also considers the relationship between the action of each task in memory that is most similar to the current action and its corresponding scenario for that task. This builds on the concept of "A is to B as C is to D", where the generated action depends not only on similar trajectories in memory, but also how these trajectories relate back to their scenario, and how the current trajectory should be related to the current scenario. By incorporating intra-task relationships, this allows the agent to better reason how the actions in memory are correlated with their own scenarios, and thereafter generate an action that is also aligned to the current scenario at hand. 

We performed evaluations using two different models: GPT-3.5 \cite{gpt3} and Llama2-13b \cite{llama2}. In Table \ref{webshop-table}, experiments with GPT-3.5 demonstrate that our method (48.0\%) achieves a higher success rate compared to previous studies such as ReAct (35.0\%), Reflexion (35.0\%), and ADaPT (43.0\%). Furthermore, our method is able to achieve a higher overall reward score (76.1\%) as compared to ReAct (61.8\%), Reflexion (61.8\%) and ADaPT (64.0\%). 

\begin{table}[ht]
 \caption{WebShop Score (\%) and Success Rate(\%).}
 \label{webshop-table}
 \vskip 0.15in
 \begin{center}
 \begin{small}
 \begin{threeparttable}
  \begin{tabular}{c | c | c | c}
  \toprule
  Method(d$_{max}$=3) & Model & Score & Success Rate \\
  \midrule
  ReAct    & GPT-3.5 & 61.8 & 35.0  \\
  Reflexion    & GPT-3.5 & 61.8 & 35.0  \\
  ADaPT   & GPT-3.5 & 64.0 & 43.0  \\
  RAP(Ours)    & GPT-3.5 & \textbf{76.1} & \textbf{48.0}  \\
  \midrule
  ReAct    & Llama2-13b & 64.6 & 31.0 \\
  RAP(Ours)    & Llama2-13b & \textbf{71.1} & \textbf{36.0} \\
  \bottomrule
  \end{tabular}
 \end{threeparttable}
\end{small}
\end{center}
\vskip -0.1in
\end{table}

\begin{figure}[t]
\hspace{-100cm}
\begin{center}
 \includegraphics[width=\linewidth-2cm]{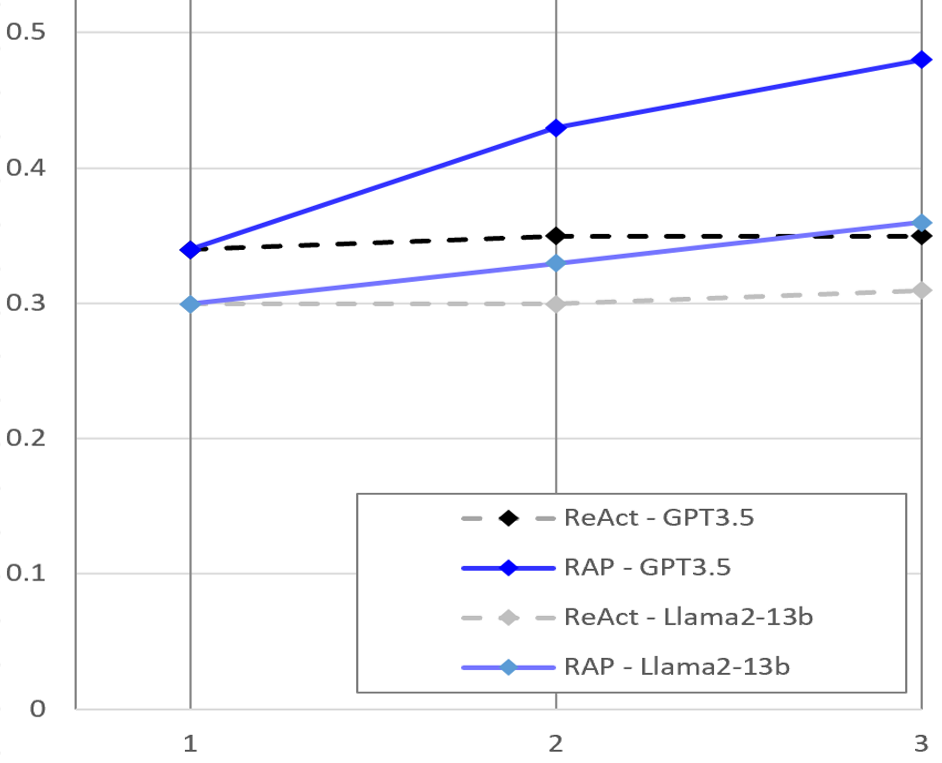}
 \vskip -0.1in
 \caption{Success Rate and Reward Score of WebShop over 3 trials}
 \label{fig:webshop_graph}
\end{center}
\vskip -0.2in
\end{figure}

Figure \ref{fig:webshop_graph} illustrates a comparison of the improvement in success rate and reward score over three trials for ReAct and RAP, with both methods using GPT-3.5. Here, it indicates a marked improvement in both success rate and reward score for our method, indicating that successful experiences from the other tasks are being effectively utilized. 

\subsection{Multimodal Environments}

\begin{figure*}[ht]
\hspace{-100cm}
\begin{center}
 \includegraphics[width=\textwidth]{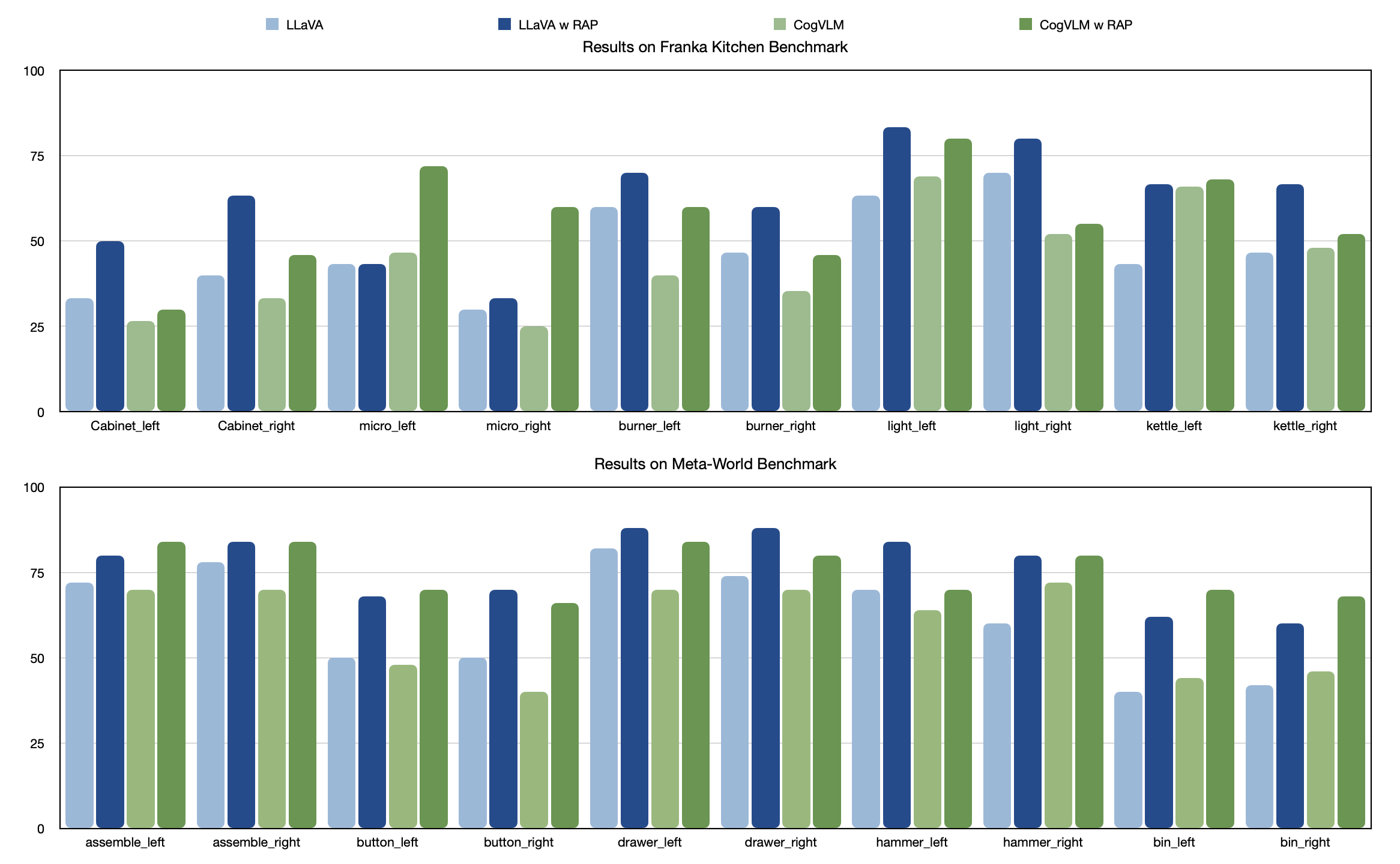}
 \vskip -0.1in
 \caption{Evaluation on Franka-Kitchen and Meta-World Benchmark. We evaluate with two vision-language models LLaVA and CogVLM both w/wo our proposed RAP method. The results demonstrate that our method notably enhances the performance of multimodal LLM agents in executing embodied tasks.}
 \label{fig:result_meta}
\end{center}
\vskip -0.2in
\end{figure*}

We evaluated our proposed technique in embodied multimodal agents on two benchmark environments: Franka Kitchen and Meta-World. These simulations offer a diverse set of household and robotic manipulation tasks requiring visual perception and physical interaction.

We constructed embodied agents using two VLM foundations - LLaVA and CogView. For each VLM, we compared task performance of the base model to a RAP-enhanced agent utilizing our memory retrieval system.

The Franka Kitchen benchmark consists of compound tasks like arranging objects and preparing meals. Meta-World provides a suite of 50 distinct robotic skills focused on fine manipulation. For both sets, the agent must plan actions based on visual observations in an interactive 3D environment.
We report quantitative results on task success rates with and without RAP augmentation. Our method allows the VLM Agents to selectively reference prior successful executions during planning. This provides vital visual context and demonstrates the benefits of memory-augmented reasoning for embodied agents.

To map the high-level plans of the VLM agents to executable environment actions, we train a policy network on 25 demonstrations for each task. We evaluate on 5 subtasks with 2 different camera views per benchmark. For each (task, view) combination we run 50 trials with different random seeds and report success rates.

Table \ref{embodied_reults} and Figure \ref{fig:result_meta} shows that RAP can significantly enhance embodied multimodal agents planning on both benchmarks.
The results offer insights into how memory can aid these models for sequential decision making and embodied tasks requiring interactive visual perception.

\begin{table}[ht]
 \caption{Average success rates on Franka Kitchen and Meta World of Vision-Languge Model  Agents w/wo RAP(\%)}
 \label{embodied_reults}
 \vskip 0.15in
 \begin{center}
 \begin{small}
 \begin{threeparttable}
  \begin{tabular}{ c | cc }
  \toprule
  Method & Franka Kitchen & Meta World \\
  \midrule
  LLaVA& 43.4 &  65.4\\
  LLaVA with RAP & \textbf{61.6} & \textbf{79.2}\\
 \midrule
  CogVLM & 44.2 & 61.5\\
  CogVLM with RAP & \textbf{56.9} &  \textbf{68.8}\\
  \bottomrule
  \end{tabular}
 \end{threeparttable}
\end{small}
\end{center}
\vskip -0.1in
\end{table}

\section{Ablation Study}

In this section, we provide additional experimental results and showcase some critical components of experiences that are required to be stored into memory.

\subsection{Evaluation across various Retrievers}

\subsubsection{ALFWorld Evaluation}

We perform evaluation on ALFWorld with GPT-3.5 by varying the Retriever, as shown in Table \ref{alfworld-table-ablation}.
The results of RAP$_{act}$ and RAP$_{obs}$ illustrate the effectiveness of switching the information used for retrieval depending on the situation.

Furthermore, we utilize visual observation provided by ALFWorld instead of textual observation, and perform an evaluation using similarity between textual retrieval key and image observation with CLIP \cite{clip}.
As a result, slightly better performance is demonstrated than when using text observation information.
This suggests that employing direct image data, rather than information converted into text, could enable more effective retrieval.

\begin{table}[ht]
 \caption{ALFWorld success rate(\%) with different retrievers. RAP$_{act}$ uses only action information, while RAP$_{obs}$ utilizes only observation information for retrieval. RAP$_{clip}$ refers to evaluations using images, rather than texts, as observations with CLIP.}
 \label{alfworld-table-ablation}
 \vskip 0.15in
 \begin{center}
 \begin{small}
 \begin{threeparttable}
  \begin{tabular}{ c | c }
  \toprule
  Method(d$_{max}$=3) & Success Rate \\
  \midrule
  ReAct & 52.2 \\
  RAP$_{act}$ & 82.1 \\
  RAP$_{obs}$ & 84.3 \\
  RAP$_{clip}$ & \textbf{86.6} \\
  RAP & \underline{85.8} \\
  \bottomrule
  \end{tabular}
 \end{threeparttable}
\end{small}
\end{center}
\vskip -0.1in
\end{table}

\subsubsection{Webshop Evaluation}
In this subsection, we illustrate the effect of varying the components extracted from each experience by the Retriever. Here, all evaluations are performed on Llama2-13b.

As shown in Table \ref{webshop-table-ablation}, through RAP$_{obs}$, the agent is able to retrieve trajectories from memory based on either actions or observations, depending on the current stage of solving the task. With the incorporation of intra-task similarity in RAP$_{intra}$, the agent is able to align the relationship between task information and the corresponding trajectories of each experience when projecting to the current task. By retrieving based on product category in RAP$_{cat}$, the agent is able to retrieve experiences that are more related to the current task. Overall, RAP takes into account these components, resulting in an overall boost of 6.5\% and 5.0\% for overall reward and success rate respectively on Llama2-13b. With these, RAP also demonstrates a boost of 14.3\% and 13.0\% for overall reward and success rate respectively on GPT-3.5 in Table \ref{webshop-table}, showcasing RAP's generalizability across different models. 

\begin{table}[ht]
 \caption{Webshop overall score (\%) and success rate(\%) with different retrievers on Llama2-13b. RAP$_{obs}$ uses additional retrieval by observations on top of action-based retrieval in RAP$_{act}$. Also, RAP$_{intra}$ and RAP$_{cat}$ uses intra-task retrieval and product-category retrieval. RAP indicates combination of RAP$_{obs}$, RAP$_{intra}$ and RAP$_{cat}$.}
 \label{webshop-table-ablation}
 \vskip 0.15in
 \begin{center}
 \begin{small}
 \begin{threeparttable}
  \begin{tabular}{ c | c | c }
  \toprule
  Method(d$_{max}$=3) & Success Rate & Overall Score\\
  \midrule
  ReAct & 31.0 & 64.6 \\
  RAP$_{act}$ & 33.0 & 68.6 \\
  RAP$_{obs}$ & 33.0 & 69.0 \\
  RAP$_{intra}$ & 34.0 & 69.3 \\
  RAP$_{cat}$ & 35.0 & 69.9 \\
  \textbf{RAP} & \textbf{36.0} & \textbf{71.1} \\
  \bottomrule
  \end{tabular}
 \end{threeparttable}
\end{small}
\end{center}
\vskip -0.1in
\end{table}

\subsection{Transfer Learning via Memory}

RAP is capable of utilizing past experiences that are stored in memory. Since the experience of solving tasks is independent of the model, the model used for evaluation does not need to match the one used for memory construction. 
Here, we illustrate a verification of transfer learning between models by using memory constructed via different models for the evaluation model.
From Section \ref{alfworld_experiment}, we use 1000 samples from training data, but no recursive trial is conducted (d$_{max} = 1$) to simply verify the effect of transfer learning.

Table \ref{alfworld-transfer} shows results of transfer learning, which indicate memory generated with GPT-3.5 is also effective in Llama2-13b. Thus, RAP allows sharing experiences across models.

\begin{table}[ht]
 \caption{ALFWorld success rate(\%) with Memory and d$_{max}$=1. Model$_{Memory}$ indicates the language model used to construct memory from the training data.}
 \label{alfworld-transfer}
 \vskip 0.15in
 \begin{center}
 \begin{small}
 \begin{threeparttable}
  \begin{tabular}{ c | c | c }
  \toprule
  Model & Model$_{Memory}$ & Success Rate \\
  \midrule
  GPT-3.5 & - & 44.0 \\
  GPT-3.5 & GPT-3.5 & 63.4 \\
  Llama2-13b & - & 20.9 \\
  Llama2-13b & GPT-3.5 & 27.6 \\
  \bottomrule
  \end{tabular}
 \end{threeparttable}
\end{small}
\end{center}
\vskip -0.1in
\end{table}

\section{Conclusion}

We propose Retrieval-Augmented Planning (RAP), which stores past experiences, extracts pertinent experiences from multimodal information such as text and images, and guides subsequent actions.
Our framework demonstrated superior performance compared to baseline methods in various LLMs and across four diverse agent and robotics benchmarks.
Through these results, our framework enables language agents to flexibly utilize past experiences in accordance with current situations, mirroring a human ability, and thereby enhancing decision-making capabilities.


\section*{Impact Statements}
This paper presents work whose goal is to advance the field of Machine Learning. There are many potential societal consequences of our work, none which we feel must be specifically highlighted here.

\bibliography{reference}
\bibliographystyle{arxiv}

\newpage
\appendix
\onecolumn

\section{Parameters}

In Webshop, all experiments were performed with a temperature setting of 0.0, due to the additional requirement of precise commands for interactions with the web application.

\begin{table}[h]
 \caption{Parameters}
 \label{parameters}
 \vskip 0.15in
 \begin{center}
  \begin{tabular}{ c c }
  \toprule
  \textbf{Experiments on ALFWorld} &  \\
  \midrule
  Text Embedding Model & sentence-transformers/all-MiniLM-L6-v2 \\
  VLM & openai/CLIP, ViT-B/32 \\
  Max steps & 50 \\
  GPT-3.5 version & gpt-3.5-turbo-instruct \\
  GPT-3.5 temperature & 0 \\
  GPT-4 version & gpt-4-0613 \\
  GPT-4 temperature & 0.5 \\
  Llama2-13b version & meta-llama/Llama-2-13b-chat-hf \\
  Llama2 temperature & default(1.0) \\
  \midrule
  \midrule
  \textbf{Experiments on Webshop} & \\
  \midrule
  Text Embedding Model & sentence-transformers/all-MiniLM-L6-v2 \\
  Max steps & 15 \\
  GPT-3.5 temperature & 0 \\
  Llama2-13b version & meta-llama/Llama-2-13b-chat-hf \\
  Llama2 temperature & 0 \\
  \bottomrule
  \end{tabular}
 \end{center}
 \vskip 0.15in
\end{table}

\section{Prompts}

\subsection{ALFWorld Prompt}
\subsubsection{Prompt for overall plan in Reasoner}
\begin{screen}
\sffamily
Here are examples.

\textbf{\{}Your task is to: put some vase on safe.

$>$ To solve the task, I need to find and take a vase, then put it on safe.\textbf{\} examples x 3}

\vskip\baselineskip
Here is the task. Please make a plan from the examples.

Your task is to: put some watch on safe.

$>$ think: To solve the task,
\end{screen}

\subsubsection{Prompt for retrieval key in Reasoner}
\begin{screen}
\sffamily
Here are examples.

think: First I need to find a spraybottle. A spraybottle is more likely to appear in cabinet (1-4), countertop (1), toilet (1), sinkbasin (1-2), garbagecan (1). I can check one by one, starting with cabinet 1.

$>$ search: spraybottle

think: Now I put the first creditcard in dresser. Next, I need to find the second creditcard. I can directly go to countertop 1.

$>$ search: creditcard

think: Now I take a pen (2). Next, I need to find a desklamp. A desklamp is more likely to appear in dresser (1), shelf (1-9), bed (1), garbagecan (1), drawer (1-10). I can check one by one, starting with dresser 1.

$>$ search: desklamp

think: Now I find a lettuce (1). Next, I need to take it.

$>$ action: take

think: Now I find a pan (1). Next, I need to take it.

$>$ action: take

think: Now I find the second saltshaker (2). Next, I need to take it.

$>$ action: take

think: Now I heat an egg (2). Next, I need to put it in/on diningtable 1.

$>$ action: put

think: Now I take a spraybottle (2). Next, I need to put it in/on toilet 1.

$>$ action: put

think: Now I take an apple (1). Next, I need to go to a microwave (1) and heat it.

$>$ action: heat

think: Now I take a bread (1). Next, I need to go to a microwave (1) and heat it.

$>$ action: heat

think: Now I take a mug (3). Next, I need to go to a fridge (1) and cool it.

$>$ action: cool

think: Now I take a potato (2). Next, I need to go to a fridge (1) and cool it.

$>$ action: cool

think: Now I find a desklamp (1). Next, I need to use it.

$>$ action: use

think: Now I find a desklamp (3). Next, I need to use it.

$>$ action: use
\vskip\baselineskip
Here is the task. Please make a plan from the examples.

think: First I need to find a watch. A watch is more likely to appear in/on cabinet (1-4), drawer (1-5). I can check one by one, starting with cabinet 1

$>$
\end{screen}

\subsubsection{Prompt for action plan in Reasoner and for action in Executor}
\begin{screen}
\sffamily
Here is the task information.

You are in the middle of a room. Looking quickly around you, you see a armchair 1, a cabinet 4, a cabinet 3, a cabinet 2, a cabinet 1, a drawer 5, a drawer 4, a drawer 3, a drawer 2, a drawer 1, a dresser 1, a garbagecan 1, a safe 1, a shelf 12, a shelf 11, a shelf 10, a shelf 9, a shelf 8, a shelf 7, a shelf 6, a shelf 5, a shelf 4, a shelf 3, a shelf 2, a shelf 1, a sidetable 1, and a sofa 1.
\vskip\baselineskip
Here are examples.

\textbf{\{}Task: put a vase in safe.

Plan: To solve the task, I need to find and take a vase, then put it in the safe.

$>$ think: First I need to find a vase.

OK.

$>$ go to shelf 1

On the shelf 1, you see nothing.

...

$>$ go to shelf 6

On the shelf 6, you see a vase 2.

$>$ think: Now I find a vase (2). Next, I need to take it

OK.

$>$ take vase 2 from shelf 6

 You pick up the vase 2 from the shelf 6.\textbf{\} examples x n}
\vskip\baselineskip
Here is the task. Please make an action from the examples.

Task: put some watch on safe.

Plan: To solve the task, I need to find and take a watch, then put it on the safe.

$>$ 
\end{screen}

\newpage

\section{ALFWorld trajectories}
\begin{itembox}[l]{\textsf{Task Description}}
\sffamily
You are in the middle of a room. Looking quickly around you, you see a bed 1, a desk 2, a desk 1, a drawer 6, a drawer 5, a drawer 4, a drawer 3, a drawer 2, a drawer 1, a garbagecan 1, a laundryhamper 1, a safe 1, a shelf 6, a shelf 5, a shelf 4, a shelf 3, a shelf 2, and a shelf 1.

\colorbox{blue!20}{Your task is to: put a mug in desk.}
\end{itembox}
\begin{minipage}[t]{0.48\columnwidth}
\centering
\begin{itembox}[l]{\textsf{ReAct}}
\sffamily

Act 1: think: To solve the task, I need to find and take a mug, then put it in desk

Obs 1: OK.

\colorbox{red!20}{\parbox{\dimexpr\linewidth-2\fboxsep}{Act 2: think: First I need to find a mug. A mug is more likely to appear in drawer (1-6), shelf (1-6), desk (1-2), garbagecan (1), laundryhamper (1), safe (1). I can check one by one, starting with drawer 1}}

Obs 2: OK.

Act 3: go to drawer 1

Obs 3: The drawer 1 is closed.

Act 4: open drawer 1

Obs 4: You open the drawer 1. The drawer 1 is open. In it, you see a pen 1.

...

Act 23: think: Hmm, it seems like the mug is not in any of the drawers. Let me check the other possible locations

Obs 23: OK.

...

\colorbox{red!20}{Fail}

\end{itembox}
\end{minipage}
\hspace{0.04\columnwidth} 
\begin{minipage}[t]{0.48\columnwidth}
\centering
\begin{itembox}[l]{\textsf{RAP}}
\sffamily

Plan: To solve the task, I need to find and take a mug, then put it in desk.

Act 1: think: First I need to find a mug

Obs 1: OK.

\colorbox{violet!20}{Retrieval key: search mug}

\colorbox{green!20}{Act 2: go to shelf 1}

Obs 2: On the shelf 1, you see a mug 2.

\colorbox{green!20}{Act 3: take mug 2 from shelf 1}

Obs 3: You pick up the mug 2 from the shelf 1.

Act 4: think: Now I find a mug (2). Next, I need to put it in/on desk 1

Obs 4: OK.

\colorbox{violet!20}{Retrieval key: action put}

Act 5: go to desk 1

Obs 5: On the desk 1, you see a bowl 1, a cd 2, a creditcard 2, a desklamp 1, a keychain 1, a laptop 2, a pen 2, and a pencil 2.

\colorbox{green!20}{Act 6: put mug 2 in/on desk 1}

Obs 6: You put the mug 2 in/on the desk 1.

\colorbox{green!20}{Success}
\end{itembox}
\end{minipage}

\newpage

\begin{itembox}[l]{\textsf{Task Description}}
\sffamily
You are in the middle of a room. Looking quickly around you, you see a cabinet 4, a cabinet 3, a cabinet 2, a cabinet 1, a countertop 1, a garbagecan 1, a handtowelholder 2, a handtowelholder 1, a sinkbasin 2, a sinkbasin 1, a toilet 1, a toiletpaperhanger 1, and a towelholder 1.

\colorbox{blue!20}{Your task is to: clean some cloth and put it in cabinet.}
\end{itembox}

\begin{minipage}[t]{0.48\columnwidth}
\centering
\begin{itembox}[l]{\textsf{ReAct}}
\sffamily

Act 1: think: To solve the task, I need to find and take a cloth, then clean it with sinkbasin, then put it in a cabinet

Obs 1: OK.

Act 2: think: First I need to find a cloth. A cloth is more likely to appear in a cabinet (1-4), countertop (1), sinkbasin (1-2), handtowelholder (1-2), towelholder (1), toiletpaperhanger (1), garbagecan (1). I can check one by one, starting with cabinet 1

Obs 2: OK.

...

Act 6: open cabinet 2

Obs 6: You open the cabinet 2. The cabinet 2 is open. In it, you see a cloth 1.

\colorbox{green!20}{Act 7: take cloth 1 from cabinet 2}

Obs 7: You pick up the cloth 1 from the cabinet 2.

Act 8: think: Now I take a cloth (1). Next, I need to go to sinkbasin (1-2) and clean it

Obs 8: OK.

Act 9: go to sinkbasin 1

Obs 9: On the sinkbasin 1, you see nothing.

Act 10: go to sinkbasin 2

Obs 10: On the sinkbasin 2, you see nothing.

\colorbox{red!20}{\parbox{\dimexpr\linewidth-2\fboxsep}{Act 11: think: I can't clean the cloth without water. I need to find a source of water}}

Obs 11: OK.

Act 12: go to toilet 1

...

\colorbox{red!20}{Fail}

\end{itembox}
\end{minipage}
\hspace{0.04\columnwidth} 
\begin{minipage}[t]{0.48\columnwidth}
\centering
\begin{itembox}[l]{\textsf{RAP}}
\sffamily
Plan: To solve the task, I need to find and take a cloth, then clean it with sinkbasin, then put it in cabinet.

Act 1: think: First I need to find a cloth

Obs 1: OK.

\colorbox{violet!20}{Retrieval key: search cloth}

Act 2: go to cabinet 1

Obs 2: The cabinet 1 is closed.

Act 3: open cabinet 1

Obs 3: You open the cabinet 1. The cabinet 1 is open. In it, you see a toiletpaper 1.

Act 4: go to cabinet 2

Obs 4: The cabinet 2 is closed.

Act 5: open cabinet 2

Obs 5: You open the cabinet 2. The cabinet 2 is open. In it, you see a cloth 1.

\colorbox{green!20}{Act 6: take cloth 1 from cabinet 2}

Obs 6: You pick up the cloth 1 from the cabinet 2.

Act 7: think: Now I take a cloth (1). Next, I need to go to sinkbasin (1) and clean it

Obs 7: OK.

\colorbox{violet!20}{Retrieval key: action clean}

Act 8: go to sinkbasin 1

Obs 8: On the sinkbasin 1, you see nothing.

\colorbox{green!20}{Act 9: clean cloth 1 with sinkbasin 1}

Obs 9: You clean the cloth 1 using the sinkbasin 1.

Act 10: think: Now I clean a cloth (1). Next, I need to put it in/on cabinet (1-6)

Obs 10: OK.

\colorbox{violet!20}{Retrieval key: action put}

Act 11: go to cabinet 3

Obs 11: The cabinet 3 is closed.

Act 12: open cabinet 3

Obs 12: You open the cabinet 3. The cabinet 3 is open. In it, you see a soapbottle 1.

\colorbox{green!20}{Act 13: put cloth 1 in/on cabinet 3}

Obs 13: You put the cloth 1 in/on the cabinet 3.

\colorbox{green!20}{Success}

\end{itembox}
\end{minipage}


\newpage
\section{Webshop trajectories}

\begin{itembox}[l]{\textsf{Task Description}}
\sffamily
WebShop 

Instruction:  

\colorbox{blue!20}{i am looking for mn4 color foundation for my sensitive skin, and price lower than 40.00 dollars}

[Search]

\end{itembox}
\begin{minipage}[t]{0.48\columnwidth}
\centering
\begin{itembox}[l]{\textsf{ReAct}}
\sffamily

Action: search[mn4 color foundation sensitive skin]

Observation: 
[Back to Search] 
Page 1 (Total results: 50) 
[Next $>$] 
[B077PR9TL4] 
Physicians Formula The Healthy Foundation with SPF 20, LW2, 1 Fl Oz 
\$7.59 
[B07VLKYJNH] 
PÜR 4-in-1 Love Your Selfie Longwear Foundation \& Concealer - Full Coverage Foundation \& Concealer Makeup For Dark Spots, Blemishes and Imperfections - Long-Lasting Liquid Makeup \& Skincare 
\$36.0 
[B09J1X9M1P] 
Neutrogena Healthy Skin Sensitive Skin Serum Foundation with Pro-Vitamin B5, Color Correcting \& Pore Minimizing Liquid Foundation \& Face Serum, Buildable Coverage, Medium 02, 1 oz 
\$14.39 

\medskip

Action: think[B077PR9TL4 and B09J1X9M1P are mn4 color foundation for sensitive skin and price lower than 40 dollars.]

Observation: OK.

\medskip

Action: click[B077PR9TL4]

Observation: 
[Back to Search] 
[$<$ Prev] 
color [1][dc1][dn3][dn4][lc1][ln3][ln4][lw2]
[mc1][mn3][mn4][mw2]
size [1 fl oz (pack of 1)][1 ounce (pack of 1)][1 fluid ounce]
Physicians Formula The Healthy Foundation with SPF 20, LW2, 1 Fl Oz 
Price: \$7.59 
Rating: N.A. 
[Description] 
[Features] 
[Reviews] 
[Attributes] 
[Buy Now] 

\medskip

\colorbox{red!20}{\parbox{\dimexpr\linewidth-2\fboxsep}{Act 11: think[For mn4 color foundation for sensitive skin, the item has options 'lw2' and '1 fluid ounce' and seems good to buy.]}}

Observation: OK.

\medskip

\colorbox{red!20}{Action: click[lw2]}

Observation: You have clicked lw2.

\medskip

Action: click[Buy Now]

Observation: Your score (min 0.0, max 1.0): 0.6666666666666666

\medskip

\colorbox{red!20}{Fail}

\end{itembox}
\end{minipage}
\hspace{0.04\columnwidth} 
\begin{minipage}[t]{0.48\columnwidth}
\centering
\begin{itembox}[l]{\textsf{RAP}}
\sffamily

Action: search[mn4 color foundation sensitive skin]

Observation: 
[Back to Search] 
Page 1 (Total results: 50) 
[Next $>$] 
[B077PR9TL4] 
Physicians Formula The Healthy Foundation with SPF 20, LW2, 1 Fl Oz 
\$7.59 
[B07VLKYJNH] 
PÜR 4-in-1 Love Your Selfie Longwear Foundation \& Concealer - Full Coverage Foundation \& Concealer Makeup For Dark Spots, Blemishes and Imperfections - Long-Lasting Liquid Makeup \& Skincare 
\$36.0 
[B09J1X9M1P] 
Neutrogena Healthy Skin Sensitive Skin Serum Foundation with Pro-Vitamin B5, Color Correcting \& Pore Minimizing Liquid Foundation \& Face Serum, Buildable Coverage, Medium 02, 1 oz 
\$14.39 

\medskip

Action: think[B077PR9TL4, B07VLKYJNH, and B09J1X9M1P are mn4 color foundation for sensitive skin and price lower than 40.00 dollars.]

Observation: OK.

\medskip

Action: click[B077PR9TL4]

Observation: 
[Back to Search] 
[$<$ Prev] 
color [1][dc1][dn3][dn4][lc1][ln3][ln4][lw2]
[mc1][mn3][mn4][mw2]
size [1 fl oz (pack of 1)][1 ounce (pack of 1)][1 fluid ounce]
Physicians Formula The Healthy Foundation with SPF 20, LW2, 1 Fl Oz 
Price: \$7.59 
Rating: N.A. 
[Description] 
[Features] 
[Reviews] 
[Attributes] 
[Buy Now] 

\medskip

Action: think[For mn4 color foundation for sensitive skin, the item has options '1', 'dc1', 'dn3', 'dn4', 'lc1', 'ln3', 'ln4', 'lw2','mc1','mn3','mn4', and'mw2' and seems good to buy.]

Observation: OK.

\medskip

\colorbox{green!20}{Action: click[mn4]}

Observation: You have clicked mn4.

\medskip

Action: click[Buy Now]

Observation: Your score (min 0.0, max 1.0): 1.0

\medskip

\colorbox{green!20}{Success}
\end{itembox}
\end{minipage}

\newpage

\begin{itembox}[l]{\textsf{Task Description}}
\sffamily
WebShop 

Instruction:  

\colorbox{blue!20}{i want capri sun pacific cooler mixed fruit naturally flavored juice drinks, and price lower than 20.00 dollars }

[Search]

\end{itembox}
\begin{minipage}[t]{0.48\columnwidth}
\centering
\begin{itembox}[l]{\textsf{ReAct}}
\sffamily

Action: search[capri sun pacific cooler mixed fruit naturally flavored juice drinks]

Observation: 
[Back to Search] 
Page 1 (Total results: 50) 
[Next $>$] 
[B000R30X2A] 
Capri Sun Pacific Cooler Mixed Fruit Naturally Flavored Juice Drink Blend (Pack of 10) (Packaging May Vary) 
\$2.55 
[B075CSWGJR] 
Capri Sun Pacific Cooler Ready-to-Drink Juice, 6 Fl Oz (Pack of 30) 
\$100.0 
[B0083GG2SC] 
Capri Sun Coolers Variety Pack Ready-to-Drink Juice (40 Pouches, 4 Boxes of 10) 
\$100.0 

\medskip

Action: think[B000R30X2A is capri sun pacific cooler mixed fruit naturally flavored juice drinks and price lower than 20.00 dollars.]

Observation: OK.

\medskip

Action: click[B000R30X2A]

Observation: 
[Back to Search] 
[$<$ Prev] 
flavor name [assorted][pacific cooler][strawberry kiwi][variety pack]
size [6 fl oz (pack of 10)][6 fl oz (pack of 30)]
Capri Sun Pacific Cooler Mixed Fruit Naturally Flavored Juice Drink Blend (Pack of 10) (Packaging May Vary) 
Price: \$2.55 
Rating: N.A. 
[Description] 
[Features] 
[Reviews] 
[Attributes] 
[Buy Now] 

\medskip

\colorbox{green!20}{\parbox{\dimexpr\linewidth-2\fboxsep}{Action: think[For capri sun pacific cooler mixed fruit naturally flavored juice drinks, the item has options 'assorted', 'pacific cooler', and '6 fl oz (pack of 10)']}}

Observation: OK.

\medskip

Action: click[assorted]

Observation: You have clicked assorted.

\medskip

Action: click[Buy Now]

Observation: Your score (min 0.0, max 1.0): 0.6666666666666666

\medskip

\colorbox{red!20}{Fail}

\end{itembox}
\end{minipage}
\hspace{0.04\columnwidth} 
\begin{minipage}[t]{0.48\columnwidth}
\centering
\begin{itembox}[l]{\textsf{RAP}}
\sffamily

Action: search[capri sun pacific cooler mixed fruit naturally flavored juice drinks]

Observation: 
[Back to Search] 
Page 1 (Total results: 50) 
[Next $>$] 
[B000R30X2A] 
Capri Sun Pacific Cooler Mixed Fruit Naturally Flavored Juice Drink Blend (Pack of 10) (Packaging May Vary) 
\$2.55 
[B075CSWGJR] 
Capri Sun Pacific Cooler Ready-to-Drink Juice, 6 Fl Oz (Pack of 30) 
\$100.0 
[B0083GG2SC] 
Capri Sun Coolers Variety Pack Ready-to-Drink Juice (40 Pouches, 4 Boxes of 10) 
\$100.0 

\medskip

Action: think[B000R30X2A is capri sun pacific cooler mixed fruit naturally flavored juice drinks and seems good to buy.]

Observation: OK.

\medskip

Action: click[B000R30X2A]

Observation: 
[Back to Search] 
[$<$ Prev] 
flavor name [assorted][pacific cooler][strawberry kiwi][variety pack]
size [6 fl oz (pack of 10)][6 fl oz (pack of 30)]
Capri Sun Pacific Cooler Mixed Fruit Naturally Flavored Juice Drink Blend (Pack of 10) (Packaging May Vary) 
Price: \$2.55 
Rating: N.A. 
[Description] 
[Features] 
[Reviews] 
[Attributes] 
[Buy Now] 

\medskip

\colorbox{green!20}{\parbox{\dimexpr\linewidth-2\fboxsep}{Action: think[For capri sun pacific cooler mixed fruit naturally flavored juice drinks, the item has options 'assorted', 'pacific cooler', and '6 fl oz (pack of 10)' and seems good to buy.]}}

Observation: OK.

\medskip

Action: click[assorted]

Observation: You have clicked assorted.

\medskip

\colorbox{green!20}{Action: click[pacific cooler]}

Observation: You have clicked pacific cooler.

\medskip

Action: click[6 fl oz (pack of 10)]

Observation: You have clicked 6 fl oz (pack of 10).

\medskip

Action: click[Buy Now]

Observation: Your score (min 0.0, max 1.0): 1.0

\medskip

\colorbox{green!20}{Success}
\end{itembox}
\end{minipage}

\newpage

\section{Multimodal Environments and Agents Details}

\subsection{Task Selection}
To evaluate RAP in multimodal environments, we focus on two benchmarks: Franka Kitchen and Meta-World. These benchmarks offer a diverse range of tasks, simulating real-world scenarios in a controlled environment, making them ideal for evaluating the performance of multimodal agents.

Franka Kitchen Benchmark simulates a kitchen environment, where the agent interacts with various kitchen appliances and items. We have selected five specific tasks that test the agent's ability to manipulate objects and controls within this environment. These tasks are listed in table \ref{table:franka_kitchen_tasks}

\begin{table}[h]
\centering
\begin{tabular}{|l|l|}
\hline
\textbf{Task No.} & \textbf{Task Description}                        \\ \hline
1.      cabinet           & Open a hinge cabinet door                           \\ \hline
2.          micro       & Open the microwave door                             \\ \hline
3.    burner             & Twist knob for bottom left burner                   \\ \hline
4.       light          & Move lever to turn on light                         \\ \hline
5.         kettle        & Move kettle from bottom to top burner               \\ \hline
\end{tabular}
\caption{Selected Tasks in Franka Kitchen Benchmark}
\label{table:franka_kitchen_tasks}
\end{table}

The Meta-World benchmark is designed to evaluate an agent's skill in more generalized object manipulation tasks. We have selected five tasks that represent a broad range of actions.

\begin{table}[h]
\centering
\begin{tabular}{|l|l|}
\hline
\textbf{Task No.} & \textbf{Meta-World Tasks}                           \\ \hline
1. assemble               & Pick up a nut and place it onto a peg                        \\ \hline
2.buttion                & Press a button                                      \\ \hline
3. drawer                & Open a drawer                \\ \hline
4. hammer                & Hammer a screw on the wall                                      \\ \hline
5. bin                & Grasp the puck from one bin and place it into another bin                   \\ \hline
\end{tabular}
\caption{Selected Tasks in Meta-World Benchmark}
\label{table:meta_world_tasks}
\end{table}

\subsection{Model Specification}
In this subsection, we detail the model specifications for our multimodal agent, which is built upon two SOTA vision-language models: LLaVA and CogVLM. These models allows for a more comprehensive understanding and interaction with multimodal environments.

\textbf{LLaVA Model}: Our agent utilizes the LLaVA-v1.5 13B model. This version of LLaVA incorporates the Vicuna-v1.5 13B as its underlying Large Language Model (LLM). The integration of Vicuna-v1.5 13B, known for its robust linguistic processing capabilities.

\textbf{CogVLM Model}: For the CogVLM model, we employ the Cog-VLM 17B version. This model boasts a significant number of parameters – 10 billion are dedicated to visual understanding, and 7 billion are focused on language processing.

In our experimental setup for the language generation of these models, we set the temperature to 0.0. This setting is chosen to prioritize precision and determinism in the generated outputs, which is crucial for the consistency and reliability of the agent's planning responses in our multimodal tasks.

\subsection{Policy Network details}
Policy Network is a crucial component of our framework designed to translate high-level action plans generated by the vision-language model into precise, low-level control actions suitable for the specific action space of the environment. Our approach utilizes a Multi-Layer Perceptron (MLP) neural network for this purpose.

In both  Franka Kitchen and Meta-World environments, the policy network's learning is facilitated through a few-shot learning approach, leveraging a limited but highly informative set of demonstration data. 

For each environment, we provide 25 expert demonstrations sourced from the D4RL dataset. These demonstrations consist of trajectories that include both observations and actions, showcasing expert-level performance in the respective tasks.

In the Franka Kitchen tasks, each demonstration is composed of 50 state-action pairs, reflecting the sequence and specifics of actions required to complete the task. Meanwhile, for the Meta-World tasks, each demonstration sample comprises 500 state-action pairs.

\end{document}